\documentclass[10pt,twocolumn,letterpaper]{article}

\usepackage{iccv}      
\iccvfinalcopy
\usepackage{times}

\usepackage{graphicx}
\usepackage{amsmath}
\usepackage{amssymb}
\usepackage{booktabs}
\usepackage{multicol}
\usepackage{multirow}
\usepackage{lipsum}
\usepackage{subcaption}
\usepackage{colortbl}
\usepackage{xcolor}
\usepackage{caption}
\usepackage{booktabs}
\usepackage{overpic}
\usepackage{array}
\usepackage{pifont}
\usepackage[toc,page]{appendix}

\usepackage[pagebackref=true,breaklinks=true,letterpaper=true,colorlinks,bookmarks=false]{hyperref}
\usepackage{cleveref}

\begin{document}

\title{SPIdepth: Strengthened Pose Information for Self-supervised Monocular Depth Estimation}

\author{Mykola Lavreniuk\\
Space Research Institute NASU-SSAU 
}
\maketitle

\ificcvfinal\thispagestyle{empty}\fi
\begin{abstract}
  Self-supervised monocular depth estimation has garnered considerable attention for its applications in autonomous driving and robotics. While recent methods have made strides in leveraging techniques like the Self Query Layer (SQL) to infer depth from motion, they often overlook the potential of strengthening pose information. In this paper, we introduce SPIdepth, a novel approach that prioritizes enhancing the pose network for improved depth estimation. Building upon the foundation laid by SQL, SPIdepth emphasizes the importance of pose information in capturing fine-grained scene structures. By enhancing the pose network's capabilities, SPIdepth achieves remarkable advancements in scene understanding and depth estimation. Experimental results on benchmark datasets such as KITTI, Cityscapes, and Make3D showcase SPIdepth's state-of-the-art performance, surpassing previous methods by significant margins. Specifically, SPIdepth tops the self-supervised KITTI benchmark. Additionally, SPIdepth achieves the lowest AbsRel (0.029), SqRel (0.069), and RMSE (1.394) on KITTI, establishing new state-of-the-art results. On Cityscapes, SPIdepth shows improvements over SQLdepth of 21.7\% in AbsRel, 36.8\% in SqRel, and 16.5\% in RMSE, even without using motion masks. On Make3D, SPIdepth in zero-shot outperforms all other models. Remarkably, SPIdepth achieves these results using only a single image for inference, surpassing even methods that utilize video sequences for inference, thus demonstrating its efficacy and efficiency in real-world applications. Our approach represents a significant leap forward in self-supervised monocular depth estimation, underscoring the importance of strengthening pose information for advancing scene understanding in real-world applications. The code and pre-trained models are publicly available at \href{https://github.com/Lavreniuk/SPIdepth}{https://github.com/Lavreniuk/SPIdepth}.
    
\end{abstract}

\section{Introduction}
\label{sec:intro}

Monocular depth estimation is a critical component in the field of computer vision, with far-reaching applications in autonomous driving and robotics \cite{Geiger2013IJRR, dudek2010computational, achtelik2009stereo}. The evolution of this field has been marked by a transition towards self-supervised methods, which aim to predict depth from a single RGB image without extensive labeled data. These methods offer a promising alternative to traditional supervised approaches, which often require costly and time-consuming data collection processes by sensors such as LiDAR \cite{wang2024SQLdepth, zhao2022monovit, peng2021excavating, watson2021temporal, feng2022disentangling}.

Recent advancements have seen the emergence of novel techniques that utilize motion cues and the Self Query Layer (SQL) to infer depth information \cite{wang2024SQLdepth}. Despite their contributions, these methods have not fully capitalized on the potential of pose estimation. Addressing this gap, we present SPIdepth, approach that prioritizes the refinement of the pose network to enhance depth estimation accuracy. By focusing on the pose network, SPIdepth captures the intricate details of scene structures more effectively, leading to significant improvements in depth prediction.

SPIdepth extends the capabilities of SQL by strengthened robust pose information, which is crucial for interpreting complex spatial relationships within a scene. Our extensive evaluations on benchmark datasets such as KITTI, Cityscapes, Make3D and demonstrate SPIdepth’s superior performance, surpassing previous self-supervised methods in both accuracy and generalization capabilities. Remarkably, SPIdepth achieves these results using only a single image for inference, outperforming methods that rely on video sequences. Specifically, SPIdepth tops the self-supervised KITTI benchmark. Additionally, SPIdepth achieves the lowest AbsRel (0.029), SqRel (0.069), and RMSE (1.394) on KITTI, establishing new state-of-the-art results. On Cityscapes, SPIdepth shows improvements over SQLdepth of 21.7\% in AbsRel, 36.8\% in SqRel, and 16.5\% in RMSE, even without using motion masks. On Make3D, SPIdepth in zero-shot outperforms all other models.

The contributions of SPIdepth are significant, establishing a new state-of-the-art in the domain of depth estimation. It underscores the importance of enhancing pose estimation within the self-supervised learning. Our findings suggest that incorporating strong pose information is essential for advancing autonomous technologies and improving scene understanding.

Our main contributions are as follows:
\begin{itemize}
    \item 
      Introducing SPIdepth, a novel self-supervised approach that significantly improves monocular depth estimation by focusing on the refinement of the pose network. This enhancement allows for more precise capture of scene structures, leading to substantial advancements in depth prediction accuracy.
    \item 
      Our self-supervised method sets a new benchmark in depth estimation, outperforming all existing methods on standard datasets like KITTI and Cityscapes using only a single image for inference, without the need for video sequences. Additionally, our approach achieves significant improvements in zero-shot performance on the Make3D dataset.
\end{itemize}

\section{Related works}

\subsection{Supervised Depth Estimation}
The field of depth estimation has been significantly advanced by the introduction of learning-based methods, with Eigen \textit{et al.} \cite{eigen2014depth} that used a multiscale convolutional neural network as well as a scale-invariant loss function.
Subsequent methods have typically fallen into two categories: regression-based approaches \cite{eigen2014depth,huynh2020guiding, zhao2021transformer} that predict continuous depth values, and classification-based approaches \cite{fu2018deep, diaz2019soft} that predict discrete depth levels.

To leverage the benefits of both methods, recent works \cite{ShariqFarooqBhat2020AdaBinsDE, johnston2020self} have proposed a combined classification-regression approach. This method involves regressing a set of depth bins and then classifying each pixel to these bins, with the final depth being a weighted combination of the bin centers.

\subsection{Diffusion Models in Vision Tasks}
Diffusion models, which are trained to reverse a forward noising process, have recently been applied to vision tasks, including depth estimation. These models generate realistic images from noise, guided by text prompts that are encoded into embeddings and influence the reverse diffusion process through cross-attention layers \cite{rombach2022stablediffusion,ho2020denoising, nichol2021improved, ramesh2022hierarchical, saharia2022photorealistic}.

The VPD approach \cite{zhao2023vpd} encodes images into latent representations and processes them through the Stable Diffusion model \cite{rombach2022stablediffusion}. Text prompts, through cross-attention, guide the reverse diffusion process, influencing the latent representations and feature maps. This method has shown that aligning text prompts with images significantly improves depth estimation performance. Different newer model further improve the accuracy of multi modal models based on Stable Diffusion \cite{lavreniuk2023evp, kondapaneni2023textimage}.

\subsection{Self-supervised Depth Estimation}

Ground truth data is not always available, prompting the development of self-supervised models that leverage either the temporal consistency found in sequences of monocular videos \cite{2017Unsupervised, godard2019digging},
or the spatial correspondence in stereo vision \cite{garg2016unsupervised, godard2017unsupervised, pillai2019superdepth}.

When only single-view inputs are available, models are trained to find coherence between the generated perspective of a reference point and the actual perspective of a related point. The initial framework SfMLearner \cite{2017Unsupervised}, was developed to learn depth estimation in conjunction with pose prediction, driven by losses based on photometric alignment. This approach has been refined through various methods, such as enhancing the robustness of image reconstruction losses \cite{gordon2019depth, shu2020feature}, introducing feature-level loss functions \cite{shu2020feature, zhan2018unsupervised}, and applying new constraints to the learning process \cite{ZhenhengYang2018UnsupervisedLO, ZhichaoYin2018GeoNetUL, ranjan2019competitive, ShengjieZhu2020TheEO, PLADENet}.

In scenarios where stereo image pairs are available, the focus shifts to deducing the disparity map, which inversely correlates with depth \cite{DanielScharstein2001ATA}. This disparity estimation is crucial as it serves as a proxy for depth in the absence of direct measurements. The disparity maps are computed by exploiting the known geometry and alignment of stereo camera setups. With stereo pairs, the disparity calculation becomes a matter of finding correspondences between the two views. Early efforts in this domain, such as the work by Garg \textit{et al.}  \cite{garg2016unsupervised}, laid the groundwork for self-supervised learning paradigms that rely on the consistency of stereo images. These methods have been progressively enhanced with additional constraints like left-right consistency checks \cite{godard2017unsupervised}.

Parallel to the pursuit of depth estimation is the broader field of unsupervised learning from video. This area explores the development of pretext tasks designed to extract versatile visual features from video data. These features are foundational for a variety of vision tasks, including object detection and semantic segmentation. Notable tasks in this domain include ego-motion estimation, tracking, ensuring temporal coherence, verifying the sequence of events, and predicting motion masks for objects. \cite{SudheendraVijayanarasimhan2017SfMNetLO} have also proposed a framework for the joint training of depth, camera motion, and scene motion from videos.

While self-supervised methods for depth estimation have advanced, they still fall short in effectively using pose data.

\begin{figure*}[t]
  \centering
   \includegraphics[width=1\linewidth]{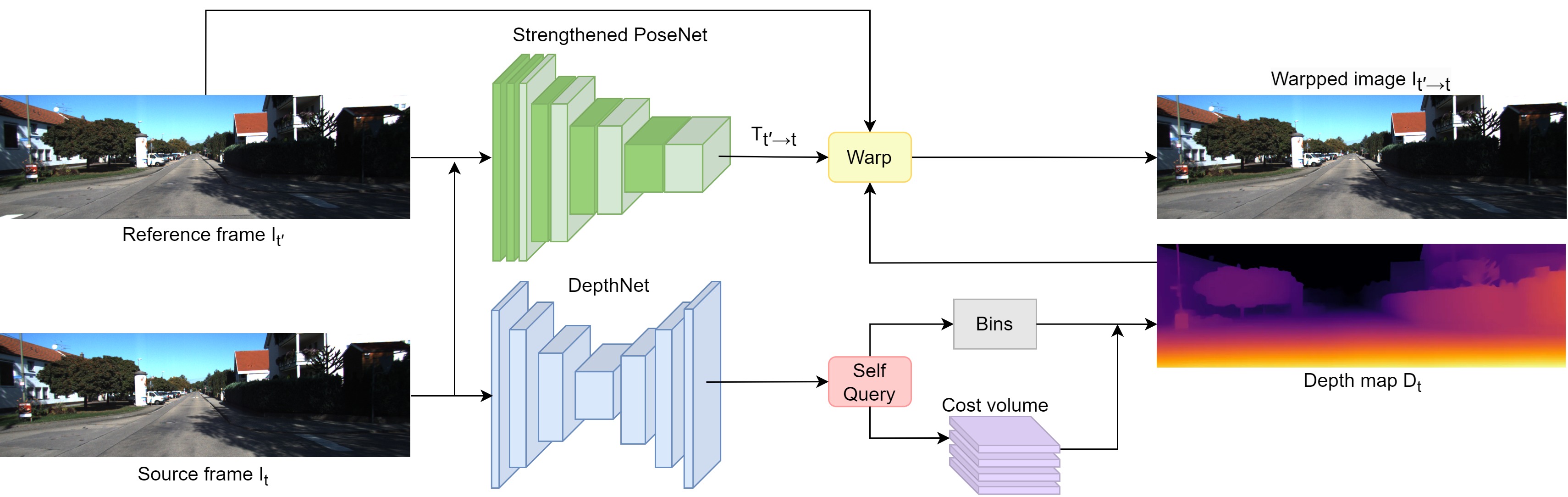}
   \hfill

   \caption{The SPIdepth architecture. An encoder-decoder extracts features from frame \(I_t\), which are then input into the Self Query Layer to obtain the depth map \(D_t\). Strengthened PoseNet predicts the relative pose between frame \(I_t\) and reference frame \(I_t'\) using a powerful pose network, needed only during training. Pixels from frame \(I_t'\) are used to reconstruct frame \(I_t\) with depth map \(D_t\) and relative pose \(T_{t' \rightarrow t}\). The loss function is based on the differences between the warped image \(I_{t' \rightarrow t}\) and the source image \(I_t\).
}
   \label{fig:our_model}
\end{figure*}

\section{Methodology}

We address the task of self-supervised monocular depth estimation, focusing on predicting depth maps from single RGB images without ground truth, akin to learning structure from motion (SfM). Our approach, SPIdepth ~\cref{fig:our_model}, introduces strengthen the pose network and enhances depth estimation accuracy. Unlike conventional methodologies that primarily focus on depth refinement, SPIdepth prioritizes improving the accuracy of the pose network to capture intricate scene structures more effectively, leading to significant advancements in depth prediction accuracy.

Our method comprises two primary components: DepthNet for depth prediction and PoseNet for relative pose estimation.

\textbf{DepthNet}: Our method employs DepthNet, a cornerstone component responsible for inferring depth maps from single RGB images. To achieve this, DepthNet utilizes a sophisticated convolutional neural network architecture, designed to extract intricate visual features from the input images. These features are subsequently processed through an encoder-decoder framework, facilitating the extraction of detailed and high-resolution visual representations denoted as \(\mathbf{S}\) with dimensions \(\mathbb{R}^{C \times h \times w}\). The integration of skip connections within the network architecture enhances the preservation of local fine-grained visual cues.

The depth estimation process could be written as:
\[
D_t = \text{DepthNet}(I_t)
\]
where \(I_t\) denotes the input RGB image.

To ensure the efficacy and accuracy of DepthNet, we leverage a state-of-the-art ConvNext as the pretrained encoder. ConvNext's ability to learn from large datasets helps DepthNet capture detailed scene structures, improving depth prediction accuracy.

\textbf{PoseNet}: PoseNet plays a crucial role in our methodology, estimating the relative pose between input and reference images for view synthesis. This estimation is essential for accurately aligning the predicted depth map with the reference image during view synthesis. To achieve robust and accurate pose estimation, PoseNet utilizes a powerful pretrained model, such as a ConvNet or Transformer. Leveraging the representations learned by the pretrained model enhances the model's ability to capture complex scene structures and geometric relationships, ultimately improving depth estimation accuracy. Given a source image \( I_t \) and a reference image \( I_{t^{\prime}} \), PoseNet predicts the relative pose \( T_{t \rightarrow t^\prime} \). The predicted depth map \( D_t \) and relative pose \( T_{t \rightarrow t^\prime} \) are then used to perform view synthesis:
\[
I_{t^\prime \rightarrow t} = I_{t^\prime}\left\langle\text{proj}\left(D_t, T_{t \rightarrow t^\prime}, K\right)\right\rangle 
\]
where \( \langle\rangle \) denotes the sampling operator and \( \text{proj} \) returns the 2D coordinates of the depths in \( D_t \) when reprojected into the camera view of \( I_{t^\prime} \).

To capture intra-geometric clues for depth estimation, we employ a Self Query Layer (SQL) \cite{wang2024SQLdepth}. The SQL builds a self-cost volume to store relative distance representations, approximating relative distances between pixels and patches. Let \( \mathbf{S} \) denote the immediate visual representations extracted by the encoder-decoder. The self-cost volume \( \mathbf{V} \) is calculated as follows:
\[
V_{i, j, k} = Q_i^T \cdot S_{j, k} 
\]
where \( Q_i \) represents the coarse-grained queries, and \( S_{j, k} \) denotes the per-pixel immediate visual representations.

We calculate depth bins by tallying latent depths within the self-cost volume \( \mathbf{V} \). These bins portray the distribution of depth values and are determined through regression using a multi-layer perceptron (MLP) to estimate depth. The process for computing the depth bins is as follows:
\[
\mathbf{b} = \text{MLP}\left(\bigoplus_{i=1}^Q \sum_{(j,k)=(1,1)}^{(h,w)}\text{softmax}(V_i)_{j,k} \cdot S_{j,k}\right)
\]
Here, \( \bigoplus \) denotes concatenation, \( Q \) represents the number of coarse-grained queries, and \( h \) and \( w \) are the height and width of the immediate visual representations.

To generate the final depth map, we combine depth estimations from coarse-grained queries using a probabilistic linear combination approach. This involves applying a plane-wise softmax operation to convert the self-cost volume \( \mathbf{V} \) into plane-wise probabilistic maps, which facilitates depth calculation for each pixel.

During training, both DepthNet and PoseNet are simultaneously optimized by minimizing the photometric reprojection error. We adopt established methodologies \cite{garg2016unsupervised, 2017Unsupervised, zhou2017unsupervised}, optimizing the loss for each pixel by selecting the per-pixel minimum over the reconstruction loss \( pe \) defined in Equation \ref{769}, where \( t^\prime \) ranges within \( (t-1, t+1) \). 
\begin{equation}
  \label{769}
L_p=\min _{t^\prime} p e\left(I_t, I_{t^\prime \rightarrow t}\right) 
\end{equation}

In real-world scenarios, stationary cameras and dynamic objects can influence depth prediction. We utilize an auto-masking strategy \cite{godard2019digging} to filter stationary pixels and low-texture regions, ensuring scalability and adaptability.

We employ the standard photometric loss combined with L1 and SSIM \cite{wang2004image} as shown in Equation \ref{loss}.
\begin{equation}
  \label{loss}
p e\left(I_a, I_b\right)=\frac{\alpha}{2}\left(1-\operatorname{SSIM}\left(I_a,
I_b\right)\right)+(1-\alpha)\left\|I_a-I_b\right\|_1
\end{equation} 
To regularize depth in textureless regions, edge-aware smooth loss is utilized.
\begin{equation}
  \label{smooth loss}
L_s=\left|\partial_x d_t^*\right|
e^{-\left|\partial_x I_t\right|}+\left|\partial_y d_t^*\right|
e^{-\left|\partial_y I_t\right|} 
\end{equation}

We apply an auto-masking strategy to filter out stationary pixels and low-texture regions consistently observed across frames.

The final training loss integrates per-pixel smooth loss and masked photometric losses, enhancing resilience and accuracy in diverse scenarios, as depicted in Equation \ref{final_loss}.
\begin{equation}
\label{final_loss}
L=\mu L_p+\lambda L_s 
\end{equation}

\begin{table*}[t]
\centering
  \resizebox{\textwidth}{!}{
  \begin{tabular}{|c|c|c|c||c|c|c|c|c|c|c|}
\hline
  Method &Train &Test &HxW &\cellcolor{red!25}$AbsRel\downarrow$ &\cellcolor{red!25}$SqRel\downarrow$ &\cellcolor{red!25}$RMSE\downarrow$ &\cellcolor{red!25}$RMESlog\downarrow$ &\cellcolor{blue!25}$\delta<1.25\uparrow$ &\cellcolor{blue!25}$\delta<1.25^2\uparrow$ &\cellcolor{blue!25}$\delta<1.25^3\uparrow$   \\ \hline
\hline
    Monodepth2 \cite{godard2019digging}  &MS &1 &1024 × 320 &0.106 &0.806 &4.630 &0.193 &0.876 &0.958 &0.980 \\
    Wang \textit{et al.} \cite{wang2020self}&M &2(-1, 0) &1024 x 320 &0.106 &0.773 &4.491 &0.185 &0.890 &0.962 &0.982 \\
    XDistill \cite{peng2021excavating} &S+Distill &1 &1024 x 320  & 0.102		& 0.698 & 4.439 	& 0.180 & 0.895	& 0.965	& 0.983 \\
    HR-Depth \cite{lyu2021hr}  &MS &1 &1024 × 320 &0.101 &0.716 &4.395 &0.179 &0.899 &0.966 &0.983 \\
    FeatDepth-MS \cite{shu2020feature} &MS &1 &1024 x 320 & 0.099		& 0.697 & 4.427 	& 0.184 & 0.889	& 0.963	& 0.982 \\
    DIFFNet \cite{zhou2021self} &M &1 &1024 x 320 & 0.097 & 0.722 & 4.345 & 0.174 & 0.907 & 0.967 & \underline{0.984} \\
    Depth Hints \cite{watson2019self} &S+Aux &1 &1024 x 320            &0.096 &0.710 &4.393 &0.185 &0.890 &0.962 &0.981   \\
    CADepth-Net \cite{yan2021channel} &MS &1 &1024 × 320 &0.096 &0.694 &4.264 &0.173 &0.908 &\underline{0.968} &\underline{0.984} \\
    EPCDepth \cite{peng2021excavating} &S+Distill &1 &1024 x 320  & 0.091 & 0.646	& 4.207 & 0.176 & 0.901	& 0.966	& 0.983 \\
    ManyDepth \cite{watson2021temporal} &M &2(-1, 0)+TTR &1024 x 320 & 0.087 & 0.685 & 4.142 & 0.167 & 0.920 & \underline{0.968} & 0.983 \\
    SQLdepth \cite{wang2024SQLdepth} &MS &1 &1024 x 320 &   \underline{0.075}  &   \underline{0.539}  & \underline{3.722}  &  \underline{0.156}  &   \underline{0.937}  &   \textbf{0.973}  &  \textbf{0.985} \\
    \rowcolor{gray!25}\textbf{SPIDepth} &MS &1 &1024 x 320 &   \textbf{0.071}  &   \textbf{0.531}  & \textbf{3.662}  &   \textbf{0.153}  &   \textbf{0.940}  &   \textbf{0.973}  &   \textbf{0.985} \\
    \hline
  \end{tabular}}
  
  \caption{
    \textbf{Performance comparison on KITTI \cite{Geiger2013IJRR} eigen benchmark.}
    In the \textit{Train} column, \textbf{S}: trained with synchronized stereo pairs, \textbf{M}: trained with monocular videos,
    \textbf{MS}: trained with monocular videos and stereo pairs,
    \textbf{Distill}: self-distillation training, \textbf{Aux}: using auxiliary information.
    In the \textit{Test} column, \textbf{1}: one single frame as input, \textbf{2(-1, 0)}: two frames (the previous and current) as input.
    The best results are in \textbf{bold}, and second best are \underline{underlined}.
    All self-supervised methods use median-scaling in \cite{eigen2015predicting} to estimate the absolute depth scale.
  }
  \label{tab:performance}
\end{table*}

\begin{table*}[ht]
\centering
  \resizebox{\textwidth}{!}{
  \begin{tabular}{|c||c|c|c|c|c|c|c|}
\hline
    Method &\cellcolor{red!25}$AbsRel\downarrow$ &\cellcolor{red!25}$SqRel\downarrow$ &\cellcolor{red!25}$RMSE\downarrow$ &\cellcolor{red!25}$RMESlog\downarrow$ &\cellcolor{blue!25}$\delta<1.25\uparrow$ &\cellcolor{blue!25}$\delta<1.25^2\uparrow$ &\cellcolor{blue!25}$\delta<1.25^3\uparrow$   \\ \hline
\hline
    BTS \cite{BTS}     &0.061 &0.261 &2.834 &0.099 &0.954 &0.992 &0.998 \\
    AdaBins \cite{ShariqFarooqBhat2020AdaBinsDE} &0.058 &0.190 &2.360 &0.088 &0.964 &0.995 &\underline{0.999} \\
    ZoeDepth \cite{bhat2023zoedepth} &0.057 &0.194 &2.290 &0.091 &0.967 &0.995 &\underline{0.999} \\
    NeWCRFs \cite{yuan2022new} &0.052 &0.155 &2.129 &0.079 &0.974 &0.997 &\underline{0.999} \\
    iDisc \cite{piccinelli2023idisc} & 0.050 & 0.148 & 2.072 & 0.076 & 0.975 & 0.997 & \underline{0.999} \\
    NDDepth \cite{shao2023nddepth} & 0.050 & 0.141 & 2.025 & 0.075 & 0.978 & \underline{0.998} & \underline{0.999} \\
    SwinV2-L 1K-MIM \cite{xie2023swinv2lmim} & 0.050 & 0.139 & 1.966 & 0.075 & 0.977 & \underline{0.998} & \textbf{1.000} \\
    GEDepth \cite{yang2023gedepth} & 0.048 & 0.142 & 2.044 & 0.076 & 0.976 & 0.997 & \underline{0.999} \\
    EVP \cite{lavreniuk2023evp} & 0.048 & 0.136 & 2.015 & 0.073 & 0.980 & \underline{0.998} & \textbf{1.000} \\
    SQLdepth \cite{wang2024SQLdepth} & 0.043  &\underline{0.105}  &\underline{1.698}  &0.064  &0.983  & \underline{0.998}  &   \underline{0.999} \\
    LightedDepth \cite{Shengjie2023LightedDepth} & \underline{0.041} &	0.107 & 1.748 &	\underline{0.059} &	\underline{0.989} &	\underline{0.998} & \underline{0.999} \\
    \rowcolor{gray!25}\textbf{SPIDepth} &\textbf{0.029}  &\textbf{0.069}  &\textbf{1.394}  &\textbf{0.048}  &\textbf{0.990}  &  \textbf{0.999}  & \textbf{1.000} \\
    \hline
  \end{tabular}}
  \caption{
    Comparison with supervised methods on KITTI \cite{Geiger2013IJRR} eigen benchmark using self-supervised pretrained and metric fine-tuned model. The best results are in \textbf{bold}, and second best are \underline{underlined}.
  }
  \label{tab: kitti eigen}
\end{table*}

\section{Results}
Our assessment of SPIDepth encompasses three widely-used datasets: KITTI, Cityscapes and Make3D, employing established evaluation metrics.

\subsection{Datasets}

\subsubsection{KITTI Dataset}
KITTI \cite{Geiger2013IJRR} provides stereo image sequences, a staple in self-supervised monocular depth estimation. We adopt the Eigen split \cite{eigen2015predicting}, using approximately 26k images for training and 697 for testing. Notably, our training procedure for SQLdepth on KITTI starts from scratch, without utilizing motion masks \cite{godard2019digging}, additional stereo pairs, or auxiliary data. During testing, we maintain a stringent regime, employing only a single frame as input, diverging from methods that exploit multiple frames for enhanced accuracy.

\subsubsection{Cityscapes Dataset}
Cityscapes \cite{cordts2016cityscapes} poses a unique challenge with its plethora of dynamic objects. To gauge SPIDepth's adaptability, we fine-tune on Cityscapes using pre-trained models from KITTI. Notably, we abstain from leveraging motion masks, a feature common among other methods, even in the presence of dynamic objects. Our performance improvements hinge solely on SPIDepth's design and generalization capacity. This approach allows us to scrutinize SPIDepth's robustness in dynamic environments. We adhere to data preprocessing practices from \cite{2017Unsupervised}, ensuring consistency by preprocessing image sequences into triples.

\subsubsection{Make3D Dataset}
Make3D \cite{saxena2008make3d} is a monocular depth estimation dataset containing 400 high-resolution RGB and low-resolution depth map pairs for training, and 134 test samples. To evaluate SPIDepth's generalization ability on unseen data, zero-shot evaluation on the Make3D test set has been performed using the SPIDepth model pre-trained on KITTI.

\subsection{KITTI Results}
We present the performance comparison of SPIDepth with several state-of-the-art self-supervised depth estimation models on the KITTI dataset, as summarized in Table \ref{tab:performance}. SPIDepth achieves superior performance compared to all other models across various evaluation metrics. Notably, it achieves the lowest values of AbsRel (0.071), SqRel (0.531), RMSE (3.662), and RMSElog (0.153), indicating its exceptional accuracy in predicting depth values.

Moving on to Table \ref{tab: kitti eigen}, we compare the performance of SPIDepth with several supervised depth estimation models on the KITTI eigen benchmark. Despite being self-supervised and metric fine-tuned, SPIDepth outperforms supervised methods across all these metrics, indicating its superior accuracy in predicting metric depth values.

Furthermore, SPIDepth surpasses LightedDepth, a model that operates on video sequences (more than one frame) and outperforms a good pre-trained models like EVP based on stable diffusion \cite{rombach2022stablediffusion}. Despite LightedDepth's advantage of using multiple frames, SPIDepth shows improvements of 0.012 (29.3\%) in AbsRel, 0.038 (34.3\%) in SqRel, 0.354 (20.3\%) in RMSE, and 0.011 (18.6\%) in RMSElog, highlighting SPIDepth's robustness and effectiveness even in challenging scenarios.

Additionally, SPIDepth demonstrates significant performance improvements over SQLdepth, a model that serves as the foundation for its development. In the self-supervised setting, SPIDepth shows improvements of 5.3\% in AbsRel,  1.5\% in SqRel, 1.6\% in RMSE, and 1.9\% in RMSElog. In the supervised setting, SPIDepth shows improvements of 32.6\% in AbsRel, 35.6\% in SqRel, 17.9\% in RMSE, and 25\% in RMSElog. These substantial improvements underscore the impact of strengthening the pose net and its information in SPIDepth.

\begin{figure*}[t]
  \centering
   \includegraphics[width=1\linewidth]{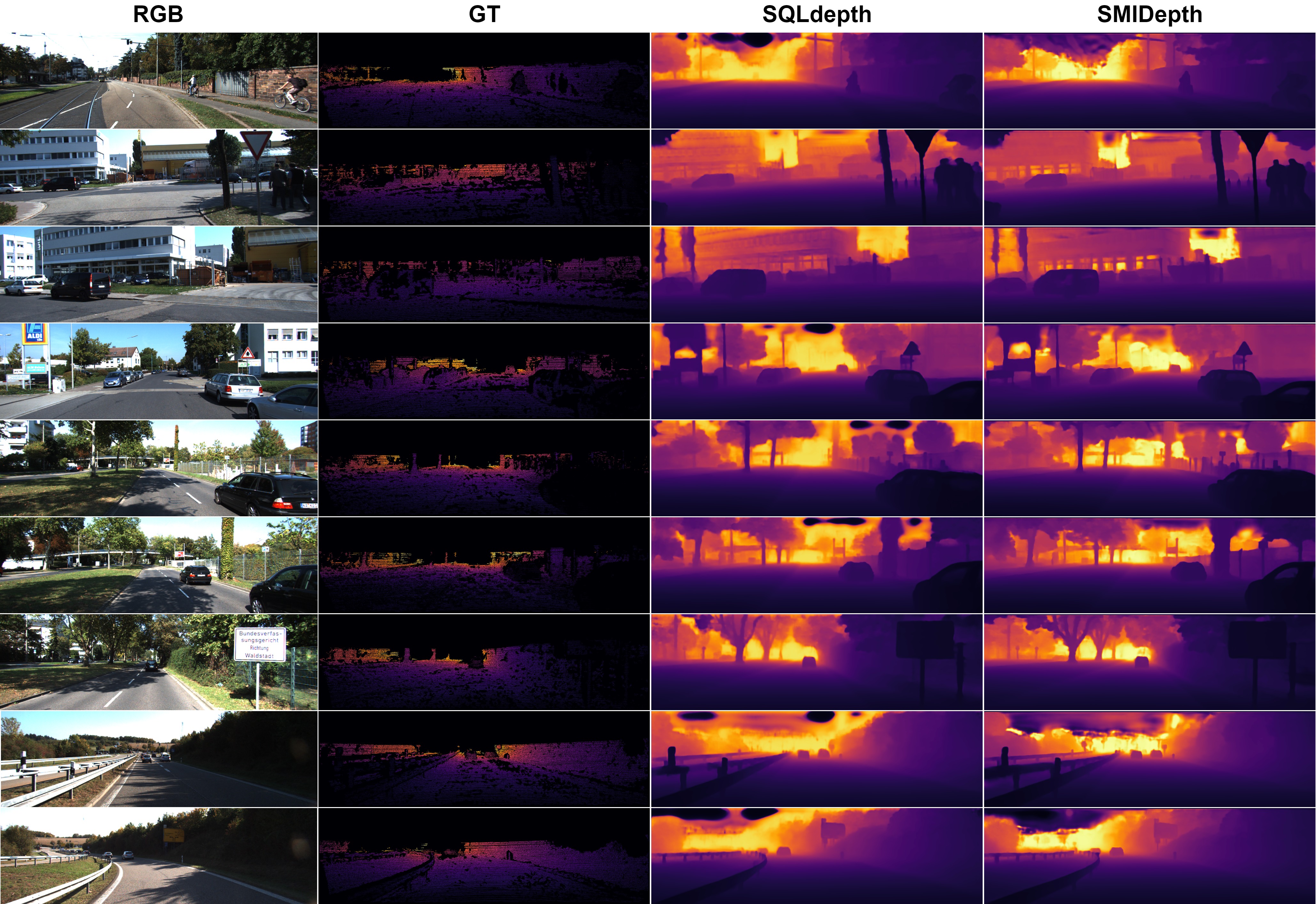}
   \hfill

   \caption{Qualitative results on the KITTI dataset. From left to right: Input RGB image, Ground Truth, SQLdepth prediction, and SPIdepth prediction.}
   \label{fig:qualitative_results}
\end{figure*}

Overall, these results underscore the effectiveness of SPIDepth in self-supervised monocular depth estimation, positioning it as a leading model in the field. Qualitative results further illustrate the superior performance of SPIDepth, as shown in Figure \ref{fig:qualitative_results}.

\begin{table*}[t]
\centering
  \resizebox{\textwidth}{!}{
  \begin{tabular}{|c|c||c|c|c|c|c|c|c|}
\hline
  Method &Train &\cellcolor{red!25}$AbsRel\downarrow$ &\cellcolor{red!25}$SqRel\downarrow$ &\cellcolor{red!25}$RMSE\downarrow$ &\cellcolor{red!25}$RMESlog\downarrow$ &\cellcolor{blue!25}$\delta<1.25\uparrow$ &\cellcolor{blue!25}$\delta<1.25^2\uparrow$ &\cellcolor{blue!25}$\delta<1.25^3\uparrow$   \\ \hline
\hline
    Pilzer \textit{et al.} \cite{AndreaPilzer2018UnsupervisedAD} &GAN, C & 0.240 & 4.264 & 8.049 & 0.334 & 0.710 & 0.871 & 0.937 \\
    Struct2Depth 2 \cite{VincentCasser2019UnsupervisedMD}  &MMask, C & 0.145 & 1.737 & 7.280 & 0.205 & 0.813 & 0.942 & 0.976 \\
    Monodepth2 \cite{godard2019digging}  &–, C & 0.129 & 1.569 & 6.876 & 0.187 & 0.849 & 0.957 & 0.983 \\
    Videos in the Wild \cite{gordon2019depth}  &MMask, C & 0.127 & 1.330 & 6.960 & 0.195 & 0.830 & 0.947 & 0.981 \\
    Li \textit{et al.} \cite{HanhanLi2020UnsupervisedMD} &MMask, C & 0.119 & 1.290 & 6.980 & 0.190 & 0.846 & 0.952 & 0.982 \\
    Lee \textit{et al.} \cite{SeokjuLee2021AttentiveAC}  &MMask, C & 0.116 & 1.213 & 6.695 & 0.186 & 0.852 & 0.951 & 0.982 \\
    ManyDepth \cite{watson2021temporal}  &MMask, C & 0.114 & 1.193 & 6.223 & 0.170 & 0.875 & 0.967 & 0.989 \\
    InstaDM \cite{SeokjuLee2021LearningMD}  &MMask, C & 0.111 & 1.158 & 6.437 & 0.182 & 0.868 & 0.961 & 0.983 \\
    SQLdepth \cite{wang2024SQLdepth} &–, K$\rightarrow$C &   0.106 & 1.173 & 6.237 & 0.163 & 0.888 & 0.972 & 0.990\\
    ProDepth \cite{woo2023ProDepth} &MMask, C & 0.095 & 0.876 & 5.531 & 0.146 & 0.908 &  0.978 & \underline{0.993}\\
    RM-Depth \cite{hui2023RM-Depth} &MMask, C & \underline{0.090} & \underline{0.825} & \underline{5.503} & \underline{0.143} & \underline{0.913} & \underline{0.980} & \underline{0.993}\\
    \rowcolor{gray!25}{\textbf{SPIDepth}} &–, K$\rightarrow$C & \textbf{0.083}  & \textbf{0.741} &   \textbf{5.205} & \textbf{0.130} & \textbf{0.931} & \textbf{0.986}  & \textbf{0.995} \\
    \hline 
  \end{tabular}}
  \caption{
  \textbf{Performance comparison on the Cityscapes \cite{cordts2016cityscapes} dataset.}
  The table presents results of models trained in a self-supervised manner on Cityscapes. \textbf{K} denotes training on KITTI, \textbf{C} denotes training on Cityscapes, and \textbf{K$\rightarrow$C} denotes models pretrained on KITTI and then fine-tuned on Cityscapes. \textbf{MMask} indicates the use of a motion mask to handle moving objects, which is crucial for training on Cityscapes, while \textbf{–} indicates no use of a motion mask. The best results are in \textbf{bold}, and second best are \underline{underlined}.
  }
  \label{tab:performance cityscapes}
\end{table*}

\subsection{Cityscapes Results}
To evaluate the generalization of SPIDepth, we conducted fine-tuning experiments in a self-supervised manner without using a motion mask on the Cityscapes dataset. Starting from a KITTI pre-trained model, we fine-tuned it on Cityscapes. The results, summarized in Table \ref{tab:performance cityscapes}, demonstrate that SPIDepth outperforms all other methods, including those that use motion masks.

Despite not using a motion mask—a technique commonly employed to handle the high proportion of moving objects in the Cityscapes dataset—SPIDepth achieves remarkable improvements over other models. Compared to SQLdepth, SPIDepth shows significant advancements: improvements of 0.023 (21.7\%) in AbsRel, 0.432 (36.8\%) in SqRel, and 1.032 (16.5\%) in RMSE.

Moreover, compared to the previous state-of-the-art model RM-Depth, which also uses motion masks, SPIDepth achieves improvements of 0.007 (7.8\%) in AbsRel, 0.084 (10.2\%) in SqRel, and 0.298 (5.4\%) in RMSE.

These results underscore SPIDepth's exceptional generalization and accuracy, achieved without the use of motion masks. This makes SPIDepth a highly robust and efficient option for depth estimation tasks. Its performance demonstrates its capability for quick deployment in new datasets, effectively addressing the challenges posed by moving objects.

\subsection{Make3D results}
To assess the generalization capacity of SPIDepth, a zero-shot evaluation was performed on the Make3D dataset \cite{saxena2008make3d} using pretrained weights from KITTI. Adhering to the evaluation settings of \cite{godard2017unsupervised, wang2024SQLdepth}, SPIDepth achieved superior results compared to other methods, including SQLdepth. Table \ref{table:Make3D_metrics} highlights these findings, showcasing the remarkable zero-shot generalization ability of the SPIDepth model.

As summarized in Table \ref{table:Make3D_metrics}, SPIDepth achieves the lowest values in all evaluation metrics, with AbsRel (0.299), SqRel (1.931), RMSE (6.672), and $log_{10}$ (0.144). These results highlight the remarkable zero-shot generalization ability of the SPIDepth model, significantly outperforming the previous best model, SQLdepth. The improvements of SPIDepth over SQLdepth are 0.007 (2.3\%) in AbsRel, 0.471 (19.6\%) in SqRel, 0.184 (2.7\%) in RMSE, and 0.007 (4.6\%) in $log_{10}$, underscoring its superior performance in challenging zero-shot scenarios.

\begin{table}[ht]
\centering
  \resizebox{0.48\textwidth}{!}{
  \begin{tabular}{|c|c||c|c|c|c|}
\hline
    Method &Type &\cellcolor{red!25}$AbsRel\downarrow$ &\cellcolor{red!25}$SqRel\downarrow$ &\cellcolor{red!25}$RMSE\downarrow$ &\cellcolor{red!25}$log_{10}\downarrow$\\ \hline
\hline
    Monodepth \cite{godard2017unsupervised}  &S &0.544 &10.94 &11.760 &0.193 \\
    Zhou \cite{zhou2017unsupervised}  &M &0.383 &5.321 &10.470 &0.478 \\
    DDVO \cite{ChaoyangWang2017LearningDF}  &M &0.387 &4.720 &8.090 &0.204 \\
    Monodepth2 \cite{godard2019digging} &M &0.322 &3.589 &7.417 &0.163 \\
    CADepthNet \cite{yan2021channel}  &M &0.312 &3.086 &7.066 &0.159 \\
    SQLdepth \cite{wang2024SQLdepth}  &M &\underline{0.306} &\underline{2.402} &\underline{6.856} &\underline{0.151} \\
    \rowcolor{gray!25}\textbf{SPIDepth}  &M &\textbf{0.299} &\textbf{1.931} &\textbf{6.672} &\textbf{0.144} \\
  \hline
  \end{tabular}}
  \caption{
  \textbf{Performance comparison on Make3D dataset \cite{saxena2008make3d}.}
    The best results are in \textbf{bold}, and second best are \underline{underlined}.
  }
  \label{table:Make3D_metrics}
\end{table}

\section{Ablation Study}
To assess the impact of Strengthened Pose Information (SPI) on depth estimation performance, we conducted an ablation study using various backbone networks, evaluated on the KITTI dataset in both self-supervised and supervised fine-tuning settings. This study compared ConvNeXt Large, ConvNeXt X-Large, and ConvNeXtV2 Huge with and without SPI, as summarized in Table \ref{table:Ablation_Study}.

Initially, we evaluated ConvNeXt Large using the standard pose net configuration, as employed in previous state-of-the-art approach - SQLdepth \cite{wang2024SQLdepth}. Without SPI, it achieved an AbsRel of 0.075 and RMSE of 3.722 in self-supervised settings. Introducing SPI improved these metrics to AbsRel 0.072 and RMSE 3.677. In supervised fine-tuning, the SPI-enhanced model showed a reduction of 0.006 (14\%) in AbsRel, and a reduction of 0.101 (5.9\%) in RMSE. ConvNeXt X-Large and ConvNeXtV2 Huge with SPI further improved performance, reaching AbsRel 0.071 and RMSE 3.662 in self-supervised settings, and AbsRel 0.029 and RMSE 1.394 in supervised fine-tuning.

While changing the backbone size provides only slight improvements in the self-supervised setting compared to the impact of SPI, it does result in more significant gains in supervised settings. These results highlight that SPI significantly enhances performance. The benefits of SPI outweigh the incremental improvements offered by larger backbones, demonstrating that SPI's impact on accuracy is more substantial than merely increasing the backbone size.

\begin{table}[ht]
\centering
  \resizebox{0.47\textwidth}{!}{
  \begin{tabular}{|c|c|c|c|c|c|}
\hline
    \multirow{2}{*}{Backbone} & \multirow{2}{*}{SPI} & \multicolumn{2}{c|}{Self-Supervised} & \multicolumn{2}{c|}{Supervised} \\
 & & \cellcolor{red!25}$AbsRel\downarrow$ & \cellcolor{red!25}$RMSE\downarrow$ & \cellcolor{red!25}$AbsRel\downarrow$ & \cellcolor{red!25}$RMSE\downarrow$ \\
\hline
    ConvNeXt Large & - & 0.075 & 3.722 & 0.043 & 1.698 \\
    ConvNeXt Large & \checkmark & 0.072 & 3.677 & 0.037 & 1.597 \\
    ConvNeXt X-Large & \checkmark & 0.071 & 3.670 & 0.034 & 1.529 \\
    \rowcolor{gray!25}{ConvNeXtV2 Huge} & \checkmark & 0.071 & 3.662 & 0.029 & 1.394 \\
  \hline
  \end{tabular}}
  \caption{
  \textbf{Ablation Study Results on KITTI Dataset.}
  The table compares the performance of different backbone networks with and without Strengthened Pose Information (SPI) in both self-supervised and supervised settings.
  }
    
  \label{table:Ablation_Study}
\end{table}

\section{Conclusion}
In summary, SPIdepth achieves significant advancements in self-supervised monocular depth estimation by enhancing the pose network during training, with no changes needed for inference. Despite adding only a minimal number of parameters compared to the depth model, SPIdepth delivers exceptional accuracy.

On the KITTI, Cityscapes, and Make3D datasets, SPIdepth sets new benchmarks in both self-supervised and fine-tuning settings, outperforming models that use multiple frames for inference. Its effectiveness in scenarios with dynamic objects and zero-shot settings demonstrates its robustness and versatility.

These results highlight SPIdepth’s potential for real-world applications, offering precise depth estimation and superior performance across diverse challenges. Its lightweight design and adaptability make it an ideal candidate for integration into various systems, enabling rapid deployment and scalable solutions in environments where accurate depth perception is crucial.

{\small
\bibliographystyle{ieee_fullname}
\bibliography{egbib}
}

\clearpage
\clearpage

\end{document}